\relax
\documentclass[letterpaper]{article} 
\usepackage{jmlr2e}
\usepackage{times,amsmath,amssymb, subfig}  
\usepackage{helvet} 
\usepackage{courier}  
\usepackage{url}  
\usepackage{graphicx} 
\usepackage{natbib}  
\usepackage{caption} 
\usepackage{xcolor}
\usepackage{amsmath, amssymb}
\usepackage{mathrsfs}

\newcommand{\name}{\texttt{UM-GNN}}
\title{Uncertainty-Matching Graph Neural Networks to \\ Defend Against Poisoning Attacks}
\author{\name Uday Shankar Shanthamallu \email ushantha@asu.edu \\
	\addr Arizona State University\\
	Tempe, AZ, USA
	\AND
	\name Jayaraman J. Thiagarajan \email jjayaram@llnl.gov \\
	\addr Lawrence Livermore National Labs\\
	Livermore, CA,USA
\AND
\name Andreas Spanias \email spanias@asu.edu \\
\addr Arizona State University\\
Tempe, AZ, USA
}

\begin{document}
	\maketitle
	
\begin{abstract}

Graph Neural Networks (GNNs), a generalization of neural networks to graph-structured data, are often implemented using message passes between entities of a graph. While GNNs are effective for node classification, link prediction and graph classification, they are vulnerable to adversarial attacks, i.e., a small perturbation to the structure can lead to a non-trivial performance degradation. In this work, we propose Uncertainty Matching GNN (\name{}), that is aimed at improving the robustness of GNN models, particularly against poisoning attacks to the graph structure, by leveraging epistemic uncertainties from the message passing framework. More specifically, we propose to build a surrogate predictor that does not directly access the graph structure, but systematically extracts reliable knowledge from a standard GNN through a novel uncertainty-matching strategy. Interestingly, this uncoupling makes \name{} immune to evasion attacks by design, and achieves significantly improved robustness against poisoning attacks. Using empirical studies with standard benchmarks and a suite of global and target attacks, we demonstrate the effectiveness of \name{}, when compared to existing baselines including the state-of-the-art robust GCN.
\end{abstract}
\section{Introduction}
\label{sec:intro}
Representation learning methods, in particular deep learning, have produced state-of-the-art results in image analysis, language modeling and more recently with graph-structured data~\cite{torng2019graph}. In particular, graph neural networks (GNNs)~\cite{Kipf2016GCNN, hamilton2017inductive} have gained prominence due to their ability to effectively leverage the inherent structure to solve challenging tasks including node classification, link prediction and graph classification~\cite{wu2020comprehensive}. 

Despite their wide-spread use, GNNs are known to be vulnerable to a variety of adversarial attacks, similar to standard deep models. In other words, a small imperceptible perturbation intentionally designed in the graph structure can lead to a non-trivial performance degradation as seen in \cite{zugner2018adversarial}. This limits their application to high-risk and safety critical domains. For example, the popular graph convolutional networks (GCN), which rely on aggregating message passes from a node's neighborhood, are not immune to poisoning attacks, wherein an attacker adds fictitious edges to the graph before the model is trained. Though there exists a vast literature on adversarial attacks on images \cite{goodfellow2014explaining, szegedy2013intriguing} and their countermeasures \cite{ren2020adversarial, chakraborty2018adversarial}, designing attack strategies for graphs is a more recent topic of research. In general, designing graph attacks poses a number of challenges: (i) the adversarial search space is discrete; (ii) nodes in the graphs are non-i.i.d., i.e., changing a link between two nodes may affect other nodes, and more importantly, (iii) lack of effective metrics to measure structural perturbations. Following the progress in graph adversarial attacks, designing defense mechanisms or building robust variants of GNNs have become critical~\cite{zhu2019robust}.

In this paper, we propose a new approach \name{} aimed at improving the robustness of GNN models, particularly against challenging poisoning attacks to the graph structure. Our approach jointly trains a standard GNN model (implemented using GCN) and a surrogate predictor, which accesses only the features, using a novel uncertainty matching strategy. Through a systematic knowledge transfer from the GNN model, the surrogate demonstrates significantly improved robustness to challenging attacks. The key contributions of this work are summarized as follows:
\begin{itemize}
    \item A novel architecture for semi-supervised learning, \name{}, that can be built upon any existing GNN model and is immune to evasion attacks by design;
    \item An uncertainty matching-based knowledge transfer strategy for achieving robustness to structural perturbations;
    \item Across a suite of global poisoning attacks, \name{} consistently outperforms existing methods including the recent Robust GCN~\cite{zhu2019robust};
    \item \name{} achieves significantly lower misclassification rate ($>50\%$ improvement) against targeted attacks.
\end{itemize}

\section{Problem Setup}
\label{sec:setup}
In this paper, we are interested in building graph neural networks that are robust to adversarial attacks on the graph structure. We represent an unweighted graph using the tuple 
$\mathrm{G} = (\mathcal{V}, \mathcal{E})$, where $\mathcal{V} = \{v_1, v_2, \cdots, v_N \}$ denotes the set of nodes with cardinality $|\mathcal{V}|$ = $N$, $\mathcal{E}$ denotes the set of edges and $\mathcal{E} \subseteq \mathcal{V} \times \mathcal{V}$. The edges in the graph may be alternately represented using an adjacency matrix $\mathbf{A} \in \mathbb{R} ^{N \times N}$. In addition, each node $v_i$ may be endowed with a $d$-dimensional node attribute vector $\mathbf{x}_i \in \mathbb{R}^d$. We use the matrix $\mathbf{X} \in \mathbb{R}^{N \times d}$ to denote the features from all nodes. We focus on a transductive learning setting, where the goal is to perform node classification. In particular, we assume that we have access to labels for a subset of nodes $\mathcal{V}_L \subset \mathcal{V}$ and we need to predict the labels for the remaining nodes ($ v \in \mathcal{V} \setminus \mathcal{V}_L$) in $\mathrm{G}$. Each node $v_i$ is associated with a label $y_i \in \mathcal{Y} = [1,\cdots,K]$.

While a variety of approaches currently exist to solve this semi-supervised learning problem, we restrict our study to the recently successful solutions based on graph neural networks (GNNs). A recurring idea in many existing GNN models is to utilize a message passing mechanism to aggregate and transform features from the neighboring nodes. Implementing a GNN hence involves designing a message function $\mathrm{P}$ and an update function $\mathrm{U}$, i.e.,
\begin{equation}
    m_i = \sum_{j \in \mathcal{N}_i} \mathrm{P} (\mathbf{h}_i, \mathbf{h}_j, e_{ij}); \quad \mathbf{h}_i= \mathrm{U} (\mathbf{h}_i, m_i),
\end{equation}where $\mathcal{N}_i$ denotes the neighborhood of a node $v_i$ and $\mathbf{h}_i$ its feature representation (in the input layer $\mathbf{h}_i = \mathbf{x}_i$). For example, in a standard graph convolutional network (GCN), 
\begin{equation}
    \mathbf{h}_i = \psi \left ( \sum_{j \in \mathcal{N}_i} \alpha_{ij}  \mathbf{h}_j \mathbf{W} \right).
    \label{eqn:gcn}
\end{equation}Here, the message computation is parameterized by $\alpha_{ij}$, which can be a symmetric normalization constant~\cite{Kipf2016GCNN} or a learnable attention weight~\cite{velickovic2018graph}. The update function $\mathrm{U}$ is parameterized using the learnable weights  $\mathbf{W}$ and applies a non-linearity $\psi$.

As discussed earlier, our goal is to defend against adversarial attacks on the graph structure. Formally, we assume that an adversary induces structural perturbations to the graph, i.e., $\hat{\mathrm{G}} = (\hat{\mathbf{A}}, \mathbf{X})$ such that $\|\mathbf{A}-\hat{\mathbf{A}}\|_0 \leq \Delta$. Here,
$\Delta$ is used to ensure that the adversarial attack is imperceptible. Note that, one can optionally also consider the setting where the features $\mathbf{X}$ are also perturbed. While different classes of attacks currently exist (see Section \ref{sec:related}), we focus on \textit{poisoning} attacks, wherein the graph is corrupted even before the predictive model is trained. This is in contrast to \textit{evasion} attacks, which assume that the model is trained on clean data and the perturbations are introduced at a later stage. We consider different popular poisoning attacks from the literature (see Section~\ref{sec:attacks}) and study the robustness of our newly proposed \name{} approach.

\section{Proposed Approach}
\label{sec:approach}
In this section, we present the proposed approach, Uncertainty Matching-GNN (\name{}), and provide details on the model training process. 

\begin{figure}
    \centering
    \includegraphics[width=0.6\linewidth]{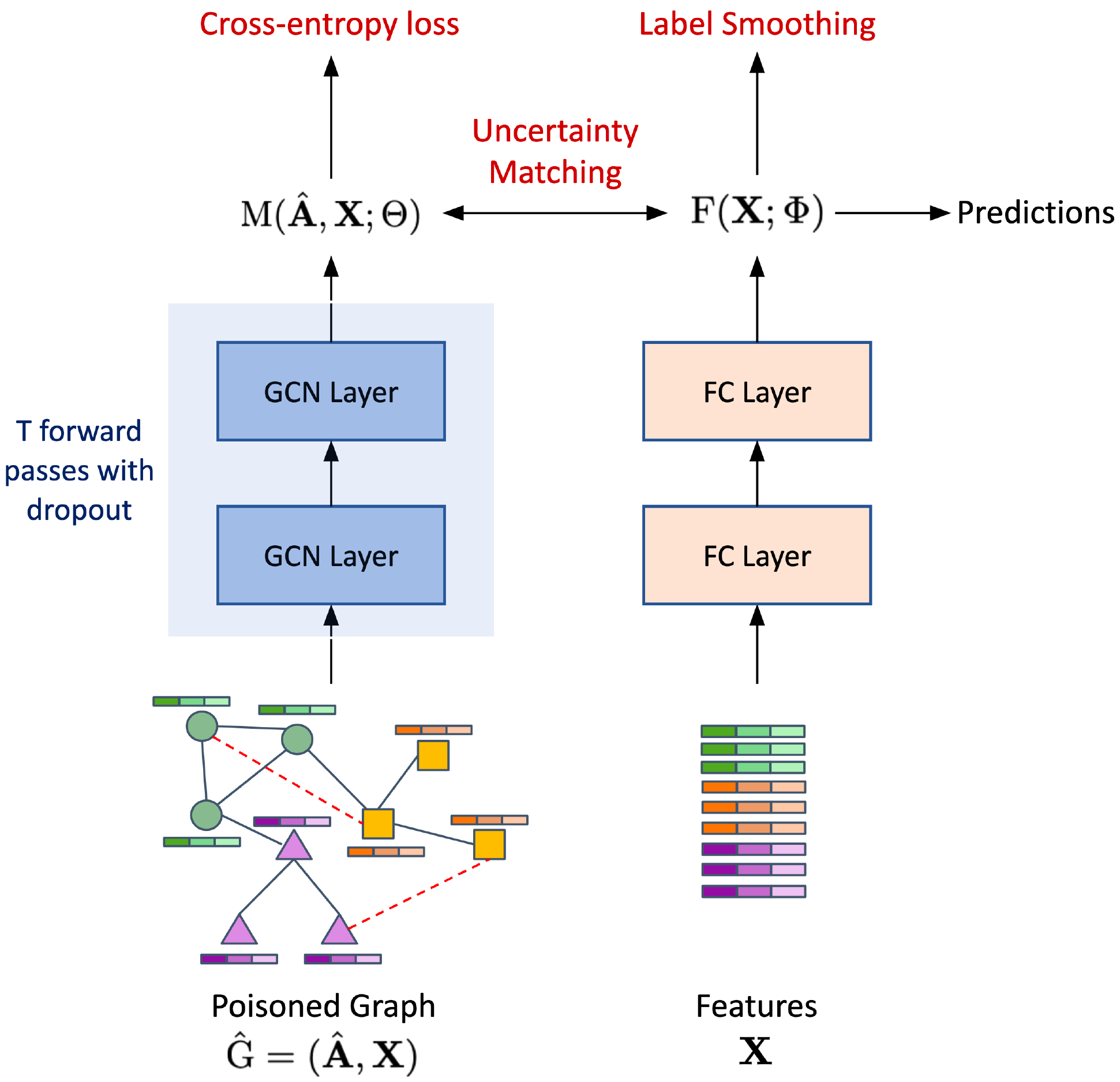}
    \caption{An illustration of the proposed \name{}, which constructs a surrogate model $\mathrm{F}$ and through an uncertainty matching strategy achieves robustness to poisoning attacks. After the model is trained, we use the surrogate model $\mathrm{F}$ to make predictions for the unlabeled nodes.}
    \label{fig:arch}
\end{figure}

While there exist very few GNN formulations for specifically defending against adversarial attacks, the recent robust GCN (RGCN) approach~\cite{zhu2019robust} has been the most effective, when compared to standard GCN and GAT models. At its core, RGCN relies on using the \textit{aleatoric} uncertainties in the graph structure to weight the neighborhood. Since there exists no \textit{a} priori knowledge about the structural uncertainties, in practice, simple priors such as the normal distribution (zero mean, unit variance) are placed on the node features and propagated through the network to estimate uncertainties at the output of each layer. Finally, a modified message passing is utilized, wherein neighboring nodes with low feature variance are emphasized during message computation to produce robust features. Despite its empirical benefits, this approach suffers from three main challenges: (i) the choice of the prior is critical to its success; (ii) since the estimated uncertainties are not calibrated, the fidelity of the uncertainty estimates themselves can be low, thus leading to only marginal improvements over GCN in practice; (iii) the model (\textit{epistemic}) uncertainties are not considered, which can impact the generalization of the inferred parameters to the test nodes. In order to alleviate these challenges, we propose \name{}, a new GNN formulation that uses an uncertainty matching-based knowledge transfer strategy for achieving robustness to graph perturbations. In contrast to RGCN, \name{} utilizes \textit{epistemic} uncertainties from the GNN, and does not require any modifications to the message passing module. As we will show in our empirical studies, our approach provides significant improvements in defending against well-known poisoning attacks.

Figure~\ref{fig:arch} provides an illustration of~\name{}, which jointly trains a GNN model $\mathrm{M}(\Theta)$ and a surrogate model $\mathrm{F}(\Phi)$ that is trained solely using the features $\mathbf{X}$ without any knowledge of the graph structure. Here $\Theta$ and $\Phi$ denote the learnable model parameters. Since we expect the graph structure to be potentially corrupted (though severity or type of corruption is unknown), the predictions from the GNN model could be unreliable due to the presence of noisy edges. We reformulate the problem of making $\mathrm{M}$ robust into systematically transferring the most reliable knowledge to the surrogate $\mathrm{F}$, so that $\mathrm{F}$ can make robust predictions. When compared to existing regularization strategies such as GraphMix~\cite{verma2019graphmix}, we neither use the (solely) feature-based model $\mathrm{F}$ to regularize the training of $\mathrm{M}$ nor are the weights shared between the networks. Instead, we build a surrogate predictor that selectively extracts the most reliable information from the ``non-robust'' $\mathrm{M}$ with the hope of being more robust to the noise in the graph structure. Interestingly, by design, the model $\mathrm{F}$ does not rely on the graph structure and hence is oblivious to evasion attacks. As showed in the figure, after training, we only use the surrogate $\mathrm{F}$ to obtain the predictions for unlabeled nodes.

\subsection{Bayesian Uncertainty Estimation}
Quantifying the prediction uncertainties in the graph neural network $\mathrm{M}$ is at the core of \name{}. We propose to utilize Bayesian Neural Networks (BNNs)~\cite{blundell2015weight}, in particular its scalable variant based on Monte Carlo dropout~\cite{srivastava2014dropout}. In general, dropout variational inference is used to estimate the epistemic uncertainties as follows: A deep network is trained with dropout and even at test time the dropout is used to generate samples from the approximate posterior through Monte Carlo sampling. Interestingly, it was showed in~\cite{gal2016dropout} that the dropout inference minimizes the KL divergence between the approximated distribution and the posterior of a deep Gaussian process. The final prediction can then be obtained by marginalizing over the posterior, using Monte Carlo integration. In our formulation, the node classification task is transductive in nature and does not require test-time inferencing. Hence, we propose to leverage the prediction uncertainties in the training loop itself. More specifically, we obtain the prediction for each node $v_i$ as
\begin{equation}
   \nonumber p(y_i = k; \mathbf{x}_i, \mathbf{A}) = \texttt{Softmax}\left(\frac{1}{T} \sum_{t=1}^T \mathrm{M}(\mathbf{x}_i, \mathbf{A}; \tilde{\Theta})\right).
    \label{eqn:mc}
\end{equation}Here we make $T$ forward passes for $\mathbf{x}_i$ with different masked weights $\tilde{\Theta}$ (using dropout inference) and compute the final prediction using a sample average. Note, we assume that the predictive model produces logits, i.e., no activation in the final prediction layer and hence compute the $\texttt{Softmax}$ of the average predictions. We then use the \textit{entropy} of the resulting prediction $p(y_i = k; \mathbf{x}_i, \mathbf{A})$ as an estimate of the model uncertainty for node $v_i$. 
\begin{align}
   \nonumber \texttt{Unc}(v_i)& = \texttt{Entropy}\bigg(p(y_i = k; \mathbf{x}_i,  \mathbf{A})\bigg)\\ &= - \sum_{k=1}^K p(y_i = k) \log p(y_i = k)
\end{align}

\subsection{Algorithm} 
We now present the algorithm to train an \name{} model given a poisoned graph $\hat{\mathrm{G}} = (\hat{\mathbf{A}}, \mathbf{X})$. As described earlier, our architecture is composed of a graph neural network $\mathrm{M}(\Theta)$ and a surrogate model $\mathrm{F}(\Phi)$ that takes only the features $\mathbf{X}$ as input. While we implement $\mathrm{M}$ using graph convolution layers as defined in  eqn.\eqref{eqn:gcn}, it can be replaced using any other message passing strategy, e.g, graph attention layers~\cite{velickovic2018graph}. Given that all datasets we consider in our study contain vector-values defined at the nodes, we implement $\mathrm{F}$ as a fully connected network. The optimization problem used to solve for the parameters $\Theta$ and $\Phi$ is given below: 
\begin{equation}
    \underset{\Theta, \Phi}{\text{minimize}} \  \mathcal{L}_{ce} + \lambda_m \mathcal{L}_{m} + \lambda_s \mathcal{L}_{s}.
\end{equation}
Here, the first term $\mathcal{L}_{ce}$ corresponds to the standard cross entropy loss over the set of labeled nodes computed using the predictions from the GNN model $\mathrm{M}$. 

\begin{figure*}
\centering
\subfloat[Citeseer with random attack]{\includegraphics[width = 0.49\linewidth]{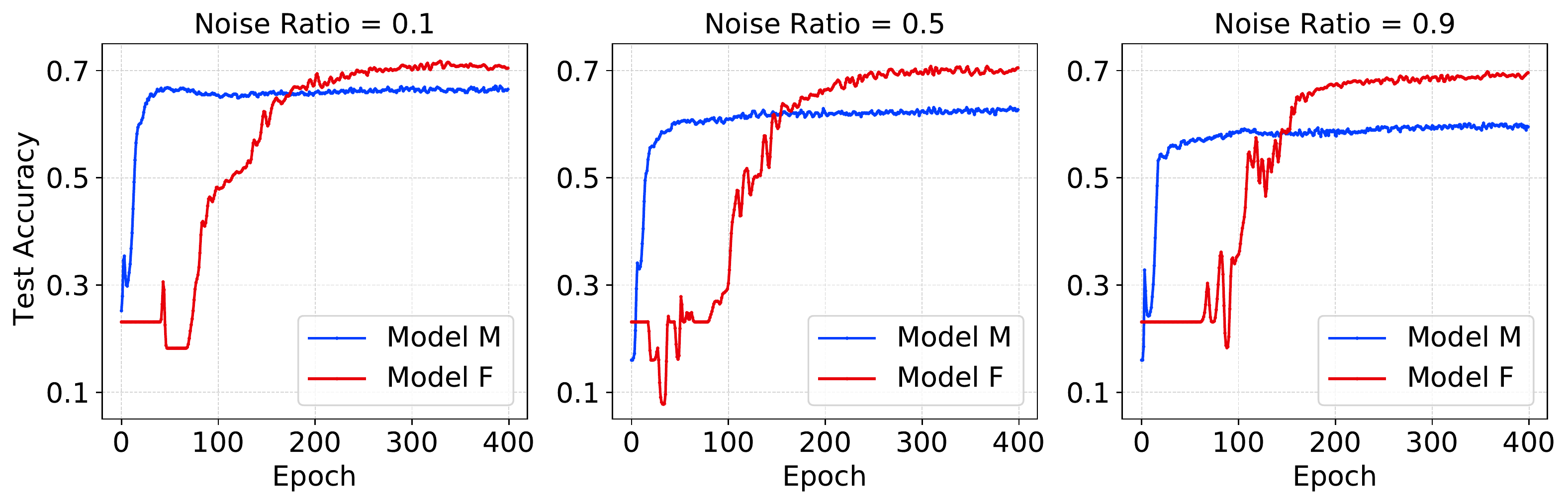}}
\subfloat[Pubmed with DICE attack]{\includegraphics[width = 0.49\linewidth]{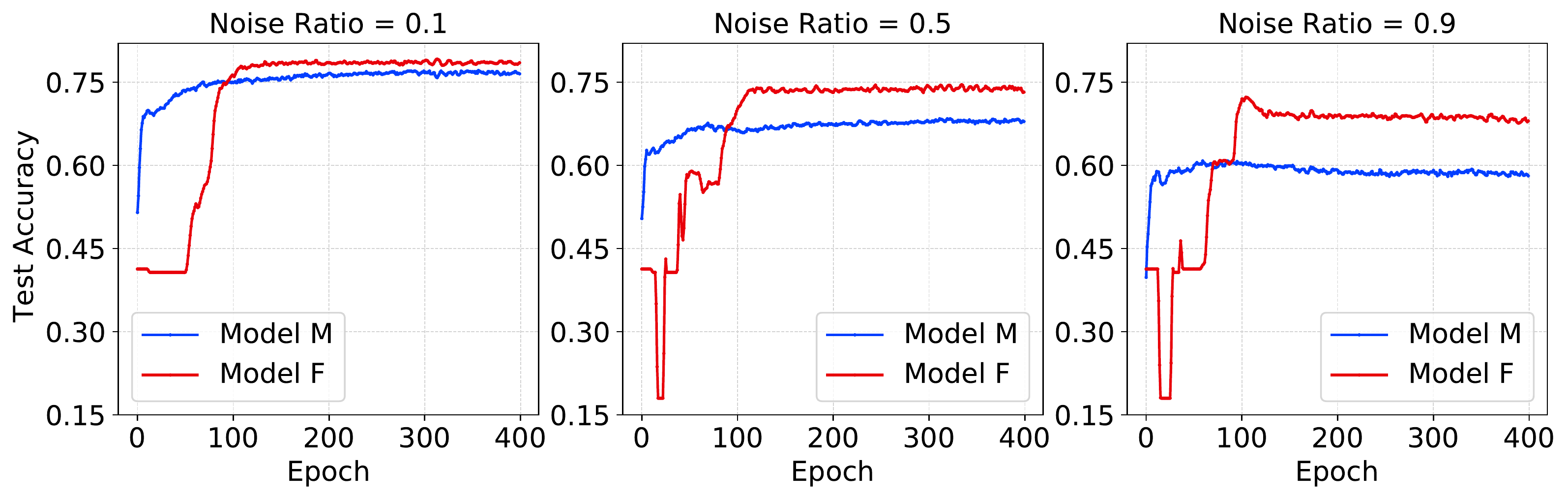}}
\caption{Illustration of the behavior of \name{} for two datasets under varying types and levels of poisoning attacks. In each case, we show the test accuracy curves across the training epochs from both the GNN and surrogate models. As the noise severity increases, the surrogate model $\mathrm{F}$ demonstrates improved robustness.}
\label{fig:example}
\end{figure*}

The second term $\mathcal{L}_{m}$ is used to align the predictions between the surrogate and GNN models so that the resulting classifiers are consistent. Directly distilling knowledge from the GNN model enables $\mathrm{F}$ to actually make meaningful predictions for the nodes, even without accessing the underlying graph structure. However, using a poisoned graph to build $\mathrm{M}$ can lead to predictions with high uncertainties. Such noisy examples may lead to unreliable gradients, thus making the knowledge transfer unstable. Hence, we propose to attenuate the influence of samples with high prediction uncertainty. We refer to this process as uncertainty matching and implement it using the KL divergence. However, this can be readily replaced using any general divergence or the Wasserstein metric. Mathematically,
\begin{equation}
    \mathcal{L}_m = \sum_{i=1}^N \beta_i \texttt{KLDiv}(\mathrm{M}(\mathbf{x}_i, \mathbf{A}; \Theta), \mathrm{F}(\mathbf{x}_i;\Phi)),
    \label{eqn:kldiv}
\end{equation}where the weight $\beta_i$s are computed as  
\begin{equation}
    \beta_i = \frac{\exp(-\alpha_i)}{\sum_j \exp(-\alpha_j)}; \text{ where } \alpha_i = \log\frac{1}{1+\texttt{Unc}(v_i)}.
\end{equation}When the prediction uncertainty for a sample is low, it is given higher attention during matching. Note that, this loss is evaluated using both labeled and unlabeled nodes, since it does not need access to the true labels. Finally, the third term $\mathcal{L}_s$ corresponds to a label smoothing regularization that attempts to match the predictions from $\mathrm{F}$ to an uniform distribution (KL divergence). This is included to safeguard the surrogate model from being misguided by the graph network, when the latter's confidences are not well-calibrated due to the poisoned graph. In all our experiments, we set $\lambda_m = 0.3$ and $\lambda_s = 0.001$. Figure~\ref{fig:example} illustrates the behavior of \name{} for two different datasets under varying levels of poisoning. As the severity of the corruption increases, the surrogate model achieves significantly higher test performance when compared to the graph-based model $\mathrm{M}$.

\section{Poisoning Attacks used for Evaluation}
While there exists a broad class of adversarial attacks that are designed to be applied during the testing phase of the model, we focus on the more challenging poisoning attacks. Poisoning attacks are intended to disrupt the model training itself by injecting carefully crafted corruptions to the training data. In particular, it is well known that they are highly effective at degrading the performance of GNNs. More importantly, existing robust modeling variants such as RGCN provide only marginal improvements over the standard GNN models, when presented with poisoned graphs. Hence, we evaluate the proposed \name{} using several widely-adopted poisoning attacks. Here, we briefly describe those attacks and provide our implementation details.


\subsubsection{Random Attack} 
This is a purely black-box attack, where the attacker has no knowledge of ground truth labels or the model information. More specifically, in this attack, new edges are randomly introduced between two nodes that were not previously connected. Though being simple, this attack is known to be effective, particularly at higher noise ratios and sparse graphs. For our experiments, we varied the ratio of noisy edges between $10\%$ and $100\%$ of the total number of edges in the original graph.

\subsubsection{DICE Attack~\cite{waniek2018hiding}} 
This is a gray-box attack where the attacker has information about the node labels but not the model parameters. This attack uses a modularity-based heuristic to Disconnect Internally (nodes from the same community) and Connect Externally (DICE) (nodes from different communities). For a given budget, an attacker randomly deletes edges that connect nodes from the same class; and adds edges between randomly chosen node pairs of samples from different classes. Similar to the random attack, we varied the perturbation ratio between $10\%$ and $100\%$ of the total number of existing edges.

\subsubsection{Meta-Gradient Attack} (Mettack) \cite{zugner2019adversarial}
This is a more challenging gray-box attack where the attacker utilizes the graph structure and labels to construct a surrogate model, which is then utilized to generate the attacks. More specifically, Mettack formulates a bi-level optimization problem of maximizing the classification error on the labeled nodes after optimizing the model parameters on the poisoned graph. In other words, the graph structure is treated as the hyper-parameter to optimize, and this is solved using standard meta-learning strategies. Since the surrogate model is also designed based on GCNs (similar architectures as our predictive model) and trained with the entire graph (transductive setting), this gray-box attack is very powerful in practice. Hence we used lower noise ratios for our experiments, i.e., between $1\%$ to $10\%$ of the total existing edges, when compared to Random and DICE attacks.

\subsubsection{Projected-Gradient Attack} (PGD) \cite{xu2019topology} PGD is a first-order topology attack that attempts to determine the minimum edge perturbations in the global structure of the graph, such that the generalization can be maximally affected. Since PGD cannot access the true model parameters, we use a surrogate GNN model to generate the attacks. Similar to Mettack, we varied the perturbation ratio between $1\%$ and $10\%$ in this case as well.

\begin{table}[t]
	\centering
\begin{tabular}{c|c|c|c|c}
\hline
\textbf{Dataset} & \textbf{\# Nodes} & \textbf{\# Edges} & \textbf{\# Features} & \textbf{\# Classes} \\ \hline \hline
Cora             & 2708              & 5278              & 1433                 & 7                   \\
Citeseer         & 3327              & 4614              & 3703                 & 6                   \\
Pubmed           & 19717             & 44325             & 500                  & 3     \\      
\hline
\end{tabular}
\caption{Summary of the three benchmark citation datasets used in our experments.}
\label{tab:datasets}
\end{table}

\subsubsection{Fast Gradient Attack} (FGA) \cite{chen2018fast} FGAs are created based on gradient information in GNNs and they belong to the category of targeted attacks. The goal of a targeted attack is to mislead the model into classifying a target node incorrectly. In FGA, the attacker adds an edge between node pairs that are characterized by largest absolute difference in their gradients. We choose FGA to show the superior performance of ~\name{} even against targeted attacks.

The implementations for Mettack, PGD and FGA were based on the publicly available DeepRobust~\cite{jin2020adversarial} library. Due to the lack of computationally efficient implementations, we could not generate these attacks on large-scale graphs such as Pubmed.

\begin{figure*}[t]
    \centering
    \includegraphics[width=.85\linewidth]{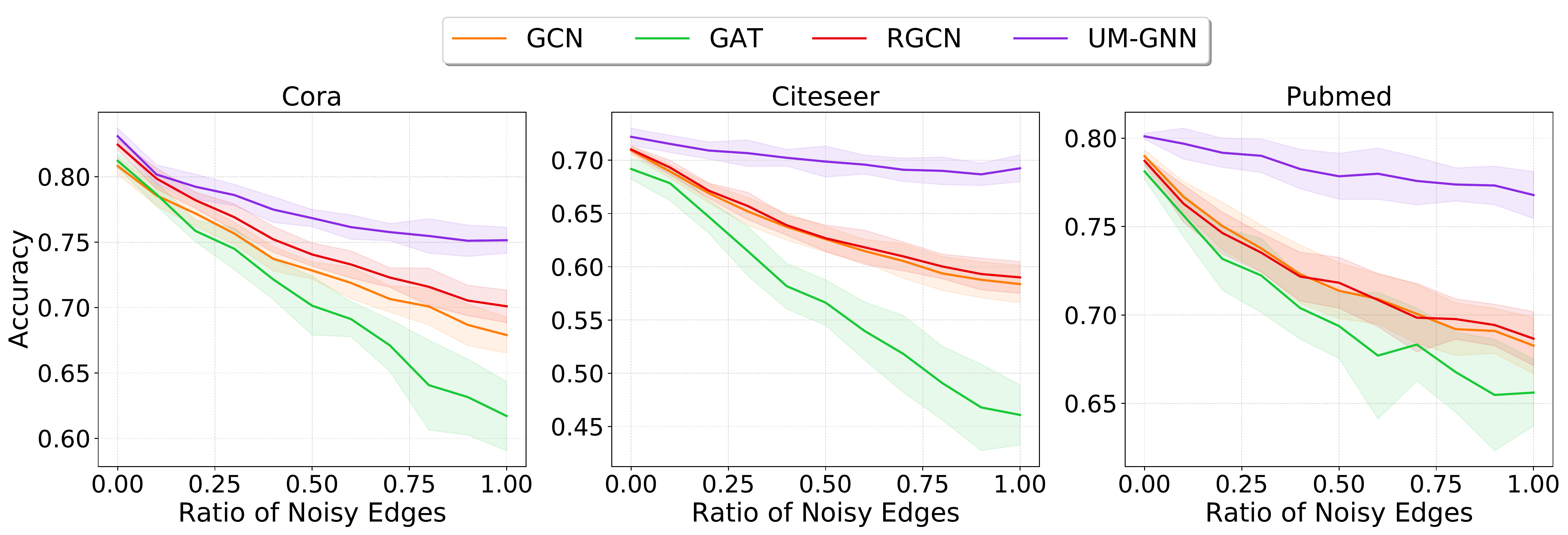}
    \caption{\textit{Random attack}: \name{} achieves robustness to random attacks, providing over $5-10\%$ improvements in the test accuracy, even when the noise ratio is $1.0$.}
    \label{fig:random}
\end{figure*}
\begin{figure*}[t]
    \centering
    \includegraphics[width=.85\linewidth]{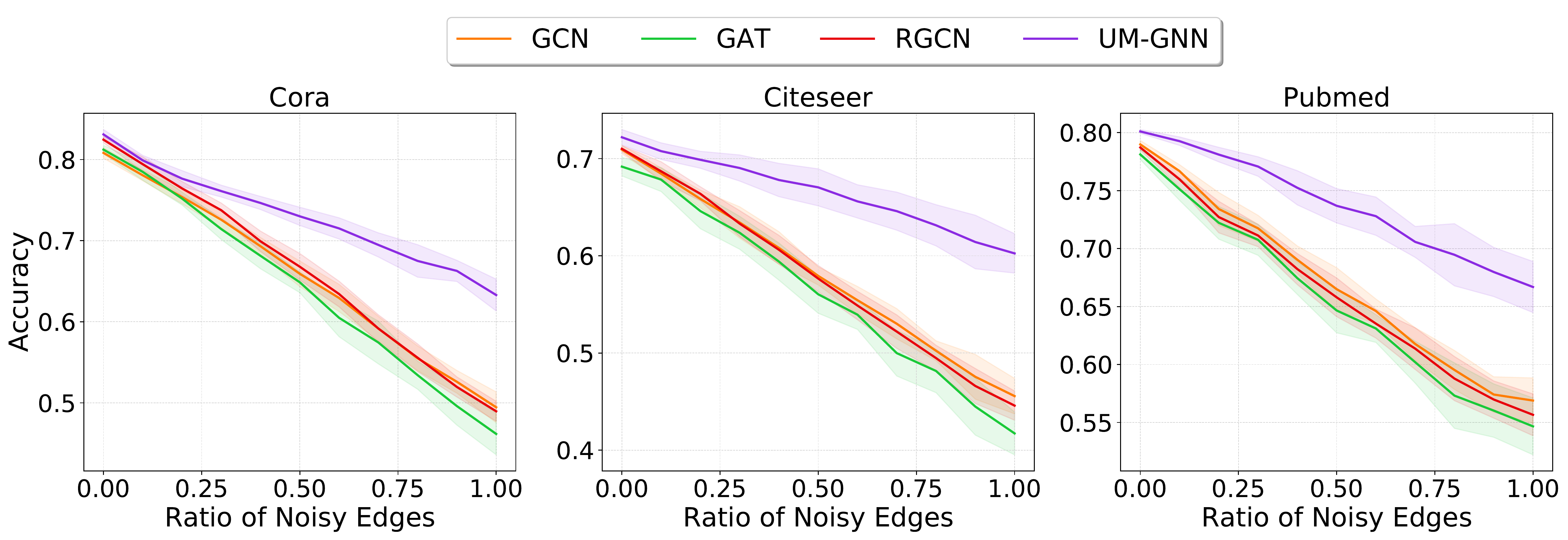}
    \caption{\textit{DICE attack}: For all datasets, \name{} is consistently more robust in this challenging scenario, where the attacker both adds and deletes edges. The performance improvement with \name{} is as high as $\approx 15\%$ (Citeseer).}
    \label{fig:dice}
\end{figure*}
\label{sec:attacks}

\section{Empirical Evaluation}
\label{sec:experiment}
In this section, we evaluate the robustness of \name{} against the graph poisoning methods discussed in the previous section. As mentioned in Section \ref{sec:attacks}, non-targeted poisoning attacks are far more challenging and pose a more realistic threat to graph-based models. 

\subsubsection{Datasets} We consider three benchmark citation networks extensively used in similar studies: Cora, Citeseer, and Pubmed \cite{sen2008collective}. The documents are represented by nodes, and citations among the documents are encoded as undirected edges. We follow the typical transductive node classification setup~\cite{Kipf2016GCNN, velickovic2018graph}, while using the standard train, test, and validation splits for our experiments (see Table \ref{tab:datasets}).

\subsubsection{Baselines}
We compare the proposed approach with three important baseline GNN models, which adopt different message passing formalisms and have been successfully used in semi-supervised node classification tasks.

\noindent \textit{GCN}: We use the GCN model, proposed by Kipf \& Welling, based on the message passing formulation in eqn. \eqref{eqn:gcn}.

\noindent \textit{GAT} \cite{velickovic2018graph}: This model uses a multi-head attention mechanism to learn the hidden representations for each node through a weighted aggregation of features in a closed neighborhood where the weights are trainable.

\noindent \textit{RGCN} \cite{zhu2019robust}: This is a recently proposed approach that explicitly enhances the robustness of GCNs. RGCN models node features as distributions as opposed to deterministic vectors in GCN and GAT models. It employs a variance-based attention mechanism to attenuate the influence of neighbors with large variance (potentially corrupted). Following~\cite{zhu2019robust}, we set hidden dimensions at $16$ and assume a diagonal covariance for each node.

For all baselines, we set the number of layers (2 layers) and other hyper-parameter settings as specified in their original papers. We set the number of hidden neurons to $16$ for both GCN and GAT baselines. In addition, we set the number of attention heads to $8$ for GAT. We implemented all the baselines and the proposed approach using the Pytorch Deep Graph Library (version 0.5.1) \cite{wang2019dgl}. In our implementation of \name{}, the GNN model $\mathrm{M}$ was designed as a $2-$layer GCN similar to the baseline and the surrogate $\mathrm{F}$ was a $3-$layer FCN with configuration $32-16-K$, where $K$ is the total number of classes.

\begin{figure}[t]
    \centering
    \includegraphics[width=.7\linewidth]{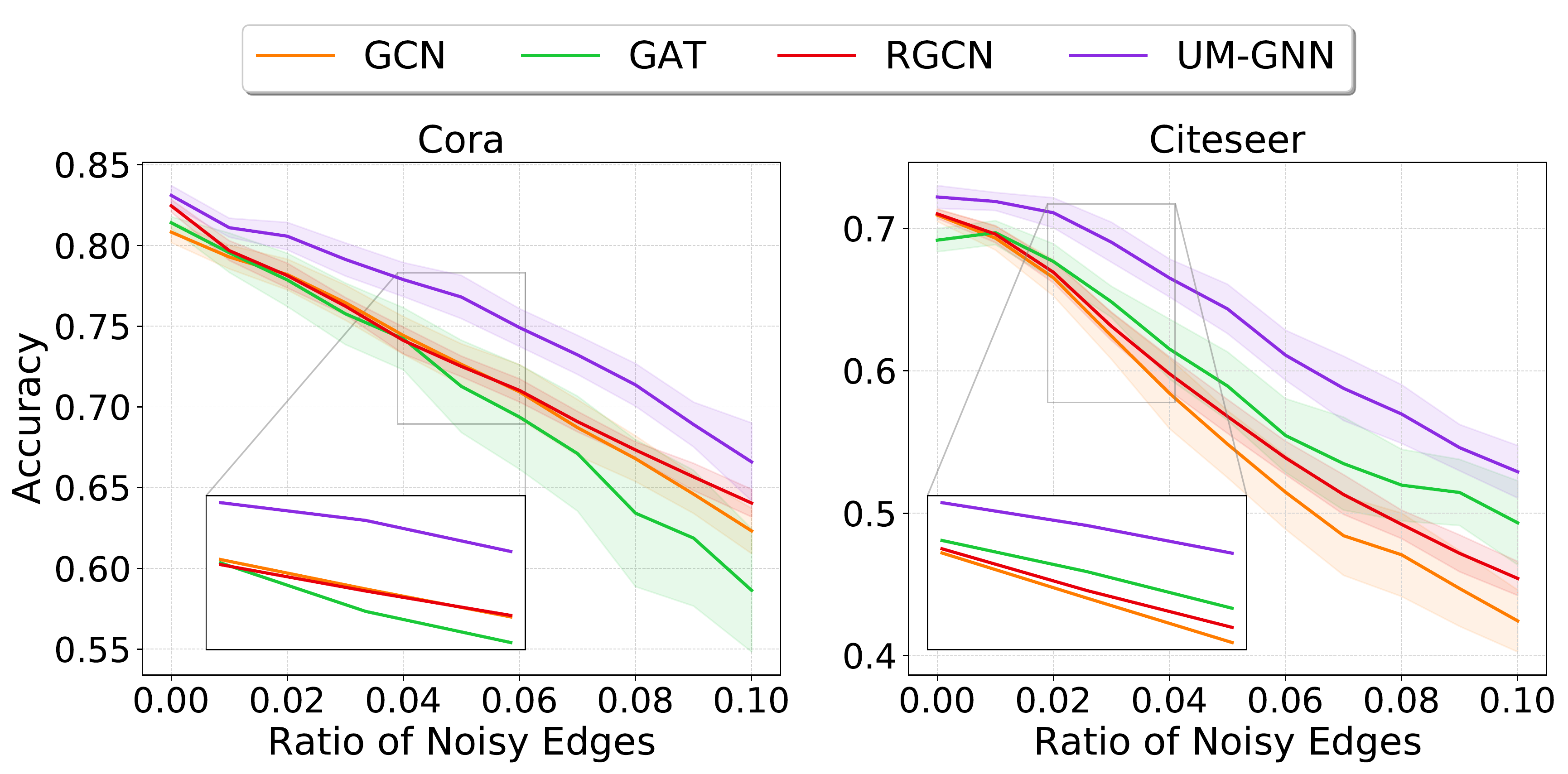}
    \caption{\textit{Mettack} - This gray-box attack is known to be highly effective at causing performance degradation in GNNs. However, \name{} consistently provides $3-5\%$ improvements in the test accuracy over the baselines.}
    \label{fig:mettack}
\end{figure}

\subsection{Results}
We evaluated the classification accuracy on the test nodes for the datasets against each of the attacks, under varying levels of perturbation. For random and DICE attacks, we varied the ratio of noisy edges to clean edges between $0.1$ and $1$. Since Mettack and PGD attacks are more powerful, we used noise ratios in the range $(0.01,0.1)$. For all the $4$ global attacks, we repeated the experiment for $20$ random trials (different corruption) for each noise ratio, and report the expected accuracies along with their standard deviations. 

\noindent\textit{(i) Random Attack}: The results for random attacks for all three datasets are shown in Figure \ref{fig:random}. As discussed earlier, RGCN provides only a marginal improvement over the vanilla GCN and GAT. However, \name{} consistently outperforms the baselines by a large margin even when the ratio of noisy edges to clean edges is high. In addition, ~\name{} has the least variance in performance compared to the baselines. In comparison, GAT appears to be the most sensitive to random structural perturbations and its low performance strongly corroborates with the findings in~\cite{zhu2019robust}.

\noindent\textit{(ii) DICE Attack}: In this challenging attack, where the attacker can both delete and add edges, all baseline methods suffer from severe performance degradation, when compared to random attacks. Surprisingly, \name{} is significantly more robust and achieves performance improvements as high as $\approx 15\%$ (Figure \ref{fig:dice}, Citeseer, noise ratio = 1.0). This clearly evidences the ability of \name{} to infer the true modular structure, even when the graph is poisoned.

\noindent\textit{(iii) Mettack Attack}: Since mettack uses a surrogate model and its parameters to generate attacks, it is one of the more challenging attacks to defend. Nevertheless, ~\name{} consistently outperforms all the baselines by a good margin, as illustrated in Figure \ref{fig:mettack}. Interestingly, under this attack, both GCN and RGCN perform poorly when compared to the GAT model. However, the large variance makes GAT unreliable in practice, particularly when the attack is severe. 

\begin{figure}[t]
    \centering
    \includegraphics[width=.7\linewidth]{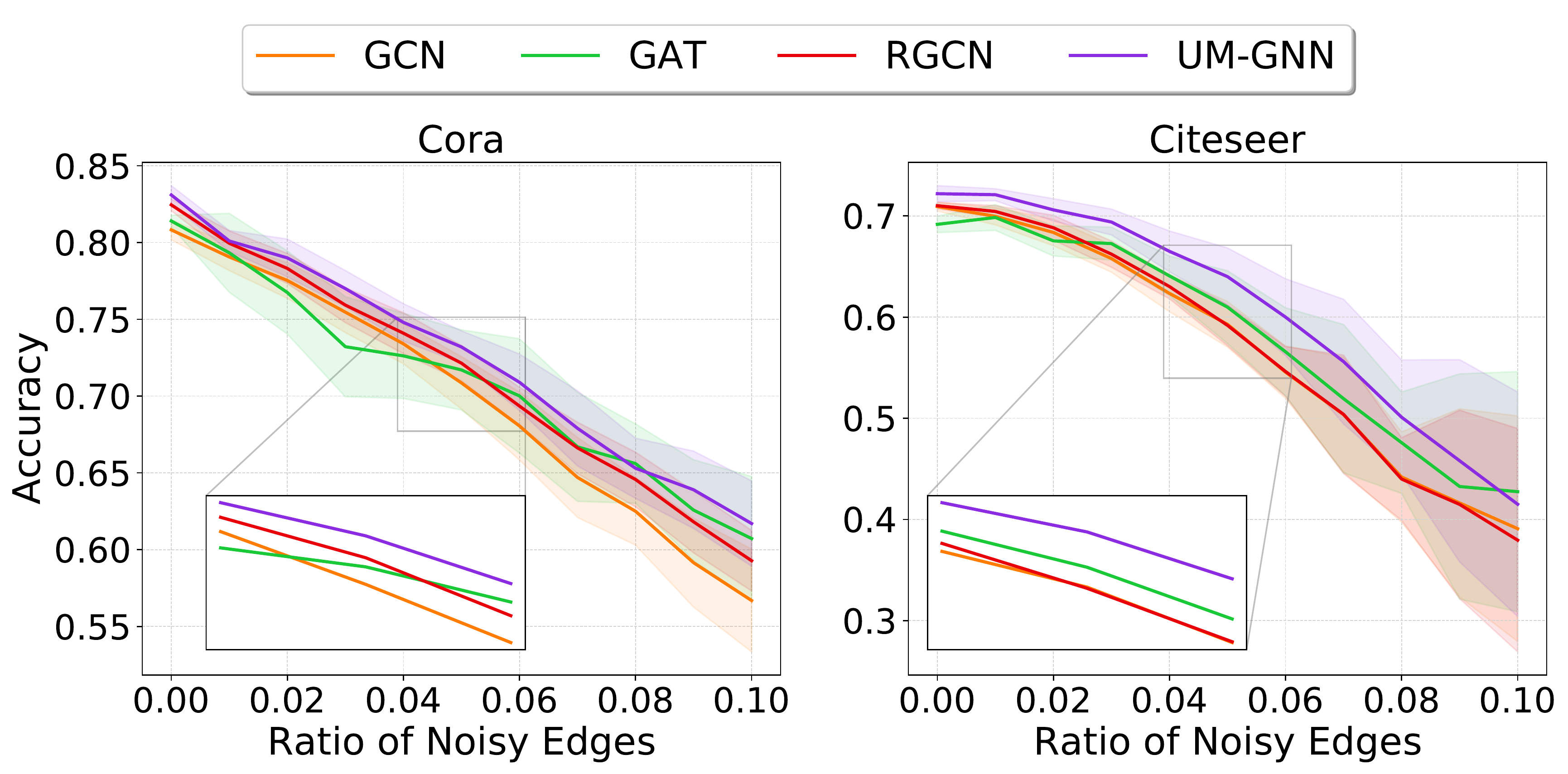}
    \caption{\textit{PGD Attack} - This is comparatively very severe, since it uses gradients from a GCN model (same architecture as $\mathrm{M}$). While the accuracy improvements are still non-trivial ($1\%-2\%$), the more interesting observation is the reduced variance of \name{} across trials.}
    \label{fig:pgd}
\end{figure}

\noindent\textit{(iv) PGD Attack}: This is comparatively the most severe, since the GCN model used to generate the attack has the same architecture as our model $\mathrm{M}$, thus in actuality making it a white-box attack. From Figure ~\ref{fig:pgd}, we observe $1\%-2\%$ improvements in mean performance over the baselines. More importantly, the lower variance of \name{} across trials makes it a suitable choice for practical scenraios.

\begin{figure*}[t]
    \centering
    \subfloat[Cora dataset]{\includegraphics[width=.9\linewidth]{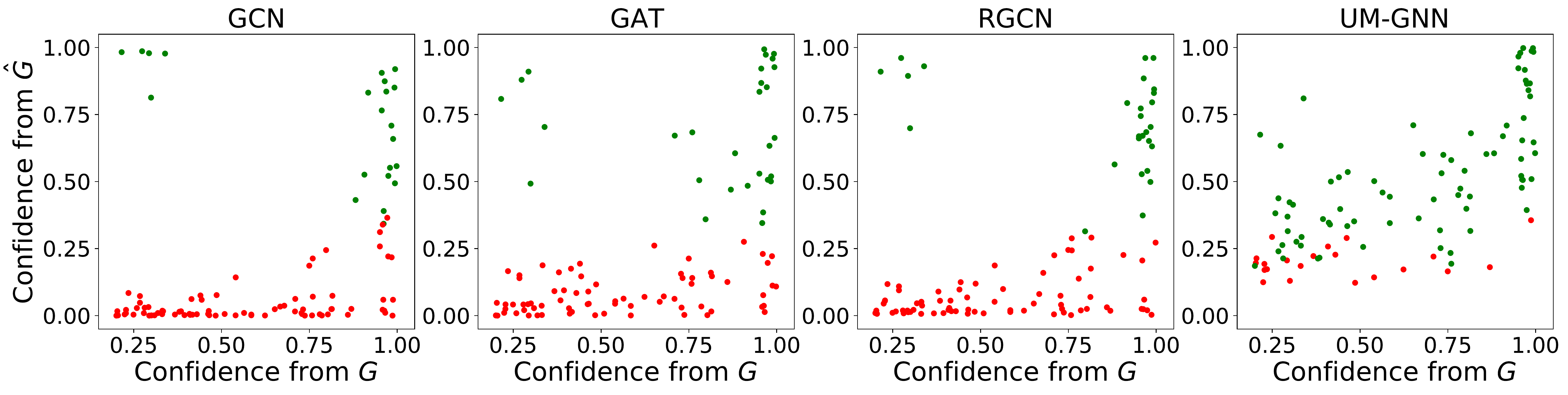}}
    \vfill
    \subfloat[Citeseer dataset]{\includegraphics[width=.9\linewidth]{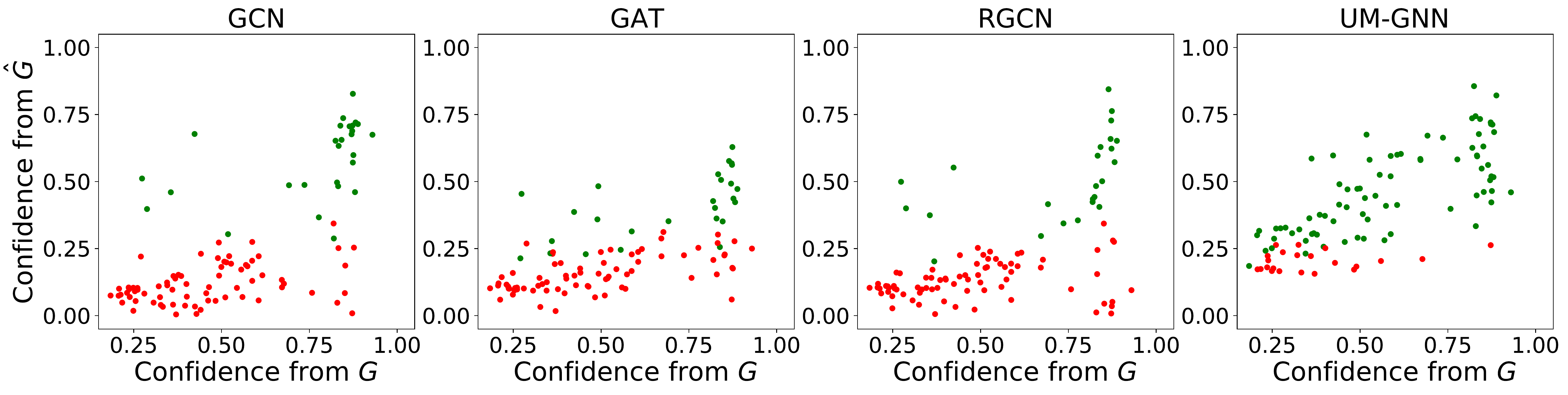}}
    \caption{Results from FGA attacks on two benchmark datasets - On the x-axis, we plot the prediction probabilities for the true class obtained using GCN on the clean graph $\mathrm{G}$. On the y-axis, we show the prediction probabilities obtained after the targeted attack. Note, for each method, we show the misclassified nodes in red and the correct predictions in green.}
    \label{fig:scatter}
\end{figure*}

\noindent\textit{(v) FGA Attack}: For this targeted attack, we selected $100$ test nodes with correct predictions in a baseline GCN as our targets. Out of the $100$ target nodes, $25$ nodes were those with the highest margin of classification, $25$ nodes were those with the lowest margin, and the remaining $50$ were chosen randomly. Further, we set the number of perturbations allowed on each target node to be equal to its degree (so that it is imperceptible). The FGA attack was generated for each target node independently, and we checked if the targeted attack was defended successfully or not, i.e., whether the targeted node was classified correctly using the poisoned graph. The overall misclassification rates for the different models are shown in Table \ref{tab:fga}. We find that \name{} provides dramatic improvements in defending against FGA attacks, through its systematic knowledge transfer between the GNN $\mathrm{M}$ and the surrogate $\mathrm{F}$. In Figure \ref{fig:scatter}, we plot the prediction probabilities for the true class (indicates a model's confidence) for all target nodes obtained using the original and poisoned graphs $\mathrm{G}$ and $\hat{\mathrm{G}}$ respectively. As it can be observed, \name{} improves the confidences considerably for all samples, while the baseline methods demonstrate vulnerability to FGA.

\section{Related Work}
\label{sec:related}

Semi-supervised learning based on graph neural networks (GNNs) enables representation learning using both the graph structure and node features~\cite{wu2020comprehensive}. While GNNs based on spectral convolutional approaches~\cite{bruna2013spectral, defferrard2016convolutional, Kipf2016GCNN} have been widely adopted, there also exists models that implement convolutions directly using spatial neighborhoods~\cite{duvenaud2015convolutional, atwood2016diffusion, hamilton2017inductive}. The vulnerability of GNNs to adversarial attacks was first studied in~\cite{zugner2018adversarial}. Since then, several graph adversarial attacks have been proposed~\cite{jin2020adversarial, sun2018adversarial}. Adversarial attacks on graphs can be broadly categorized as follows: 

\noindent(i) \textit{Attacker knowledge}: based on the level of access an attacker has to the model internals, namely white-box~\cite{xu2019topology, wu2019adversarial}, gray-box~\cite{zugner2018adversarial, zugner2019adversarial} and black-box attacks~\cite{bojchevski2019adversarial}.

\noindent(ii) \textit{Attacker capability}: based on whether the attacker perturbs the graph before \cite{liu2019unified} or after ~\cite{dai2018adversarial} the model is trained.

\noindent(iii) \textit{Attack strategy}: based on whether the attacker corrupts the graph structure or adds perturbations to the node features. While structural perturbations can be induced by deleting  edges, adding new edges or re-wiring existing edges; new nodes could also be injected into the graph~\cite{shanthamallu2020reg}.

\noindent(iv)\textit{Attacker's goal}: based on whether the attacker is aimed at degrading the model's overall performance~\cite{waniek2018hiding} or targeting specific nodes either directly or indirectly for their misclassification~\cite{chen2018fast}.

\begin{table}[t]
\renewcommand{\arraystretch}{1.0}
\centering
\begin{tabular}{c|c|c}
\hline
\textbf{Model} & \textbf{Cora} & \textbf{Citeseer} \\ \hline \hline
GCN            & 0.78          & 0.73              \\ \hline
GAT            & 0.71          & 0.74              \\ \hline
RGCN           & 0.73          & 0.76              \\ \hline
\textbf{~\name{}}        & \textbf{0.21}         & \textbf{0.23}              \\ \hline
\end{tabular}
\caption{Misclassification rates from $100$ target nodes with FGA attack. A lower value implies improved robustness.}
\label{tab:fga}
\end{table}

\paragraph{Graph Adversarial Defense}
As graph adversarial attacks continue to be studied, efforts aimed at designing suitable defense strategies have emerged recently. For example, Feng \textit{et al.} adapted the conventional adversarial training approach to the case of graphs in order to make GNNs more robust~\cite{goodfellow2014explaining, feng2019graph}. On the other hand, methods that rely on graph pre-processing have also been proposed -- for example, in~\cite{wu2019adversarial}, edges with low Jaccard similarity between the  constituent nodes were removed prior to training a GNN. Similarly, in~\cite{jin2019power}, explicit graph smoothing was performed by training on a family of graphs to defend against evasion attacks. Entezari \textit{et al.} obtained a low rank approximation of the given graph and showed that it can defend against specific types of graph attack~\cite{zugner2018adversarial}. Recently, Zhu \textit{et al.}~\cite{zhu2019robust} introduced a robust variant of GCN based on a variance-weighted attention mechanism, and showed it to be effective against different types of attacks, when compared to standard GCN and GAT models.



\section{Conclusions}
\label{sec:conclusion}
In this work, we presented \name{} an uncertainty matching-based architecture to explicitly enhance the robustness of GNN models. \name{} utilizes epistemic uncertainties from a standard GNN $\mathrm{M}$ and does not require any modifications to the message passing module. Consequently, our architecture is agnostic to the choice of GNN to implement $\mathrm{M}$. By design, the surrogate model $\mathrm{F}$ does not directly access the graph structure and hence is immune to evasion-style attacks. Our empirical studies clearly evidenced the effectiveness of \name{} in defending against several graph poisoning attacks, thereby outperforming existing baselines. Furthermore, we showed dramatic improvements on defense against targeted attacks (FGA). Future work includes studying the performance bounds of \name{} and developing extensions for inductive learning settings.

 \section{Acknowledgements}
This work was performed under the auspices of the U.S. Department of Energy by Lawrence Livermore National Laboratory under Contract DE-AC52-07NA27344. 

\bibliography{refs}


\end{document}